**Title Page**

**Original Article**

**Development of a European Union Time-Indexed Reference Dataset for Assessing the Performance of Signal Detection Methods in Pharmacovigilance using a Large Language Model**

Maria Kefala[(1)]*, Jeffery L. Painter[(2)]*, Syed Tauhid Bukhari[(1)]*, Maurizio Sessa[(1)]

(1) Department of Drug Design and Pharmacology, University of Copenhagen, Copenhagen, Denmark.

(2) GSK, Durham, NC, USA.

* These authors contributed equally as first authors to this work

**Metrics**

*Manuscript*: 5258 words

*Abstract*: 296 words

*References*: 22

*Figures and tables*: 4 figures and 2 tables

**Corresponding author**:

Maurizio Sessa,

Department of Drug Design and Pharmacology,

University of Copenhagen,

Jagtvej 160 Copenhagen 2100, Denmark

Email: maurizio.sessa@sund.ku.dk

**Abstract**

**Background:**

The identification of optimal signal detection methods is hindered by the lack of reliable reference datasets. Existing datasets do not capture when adverse events (AEs) are officially recognized by regulatory authorities, preventing restriction of analyses to pre-confirmation periods and limiting evaluation of early detection performance. This study addresses this gap by developing a time-indexed reference dataset for the European Union (EU), incorporating the timing of AEs inclusion in product labels along with regulatory metadata. This dataset is the EU analogue of the United States (US)-based PVLens initiative.

**Methods:**

Current and historical Summaries of Product Characteristics (SmPCs) for all centrally authorized products (i.e., 1,513) were retrieved from the EU Union Register of Medicinal Products (data lock: 15 December 2025). Section 4.8 ("Undesirable effects") was extracted and processed using DeepSeek V3 to identify AEs. Regulatory metadata, including labelling changes, were programmatically extracted. Time indexing was based on the date of AE inclusion in the SmPC.

**Results:**

The database includes 17,763 SmPC versions spanning 1995–2025, comprising 125,026 drug–AE associations. The processed database (i.e., time-indexed reference dataset) with only active Centrally Authorized Products (CAPs) had 1,479 medicinal products and a total of 110,823 drug-event associations. Most AEs were identified in pre-marketing (74.5%) versus post-marketing (25.5%), with similar System Organ Classes (SOC) distributions. Safety updates increased over time, peaking around 2012. AEs involved SOCs gastrointestinal, skin, and nervous system disorders were most frequently reported in the SmPCs. While most AEs were shared across products, 21.7% were product specific. Drugs had a median of 48 AEs across 14 SOCs.

**Conclusions:**

The proposed database and processed database address a critical gap in pharmacovigilance by incorporating temporal information on AEs recognition for the EU. They support more accurate assessment of signal detection performance and facilitate methodological comparisons across diverse analytical approaches.

# 1. Introduction

Current quantitative approaches for signal detection generate a substantial proportion of false-positive statistical alerts (1-3). Statistical alerts require validation by regulatory agencies which is a time- and resource-demanding procedure (4). The development of more effective signal detection methods is currently hindered by the lack of reliable reference datasets, which are currently limited in scope, size, or are outdated (5, 6). Reference datasets are essential for evaluating the performance of signal detection methods because they include drug–adverse event (AE) pairs with well-established causal relationships (i.e., true positives) that should be identified by signal detection algorithms (5). Another major, often overlooked limitation of currently available reference datasets is the lack of information on the timing of the first regulatory confirmation of drug–AE pairs. In signal detection analyses, only data preceding the regulatory confirmation of a drug–AE pair should be considered, as including data generated after regulatory validation would measure the method's ability to confirm already known associations rather than its ability to detect new safety signals.

Within the European Union (EU) regulatory framework for Centrally Authorized Products (CAPs), the continuously updated Summary of Product Characteristics (SmPC) documents and all their historical versions are archived in the European Commission's Union Register of Medicinal Products (7). This archive makes it possible to determine the time at which a drug–AE pair was first introduced in the SmPC, representing the regulatory confirmation of the association. However, to date, no documented effort has been made to systematically extract and compile this information from the European Commission's Union Register of Medicinal Products to create a publicly available, time-indexed reference dataset of drug–AE pairs for CAPs in the EU. Recent work has explored the use of large language models (LLMs) for extraction of safety information from regulatory labeling, particularly in settings where adverse event content is embedded in semi-structured tables or irregular textual formats. These approaches demonstrate improved flexibility in recovering adverse event expressions compared with rule-based mapping processes, but raise important considerations regarding normalization, reproducibility, and alignment with controlled terminologies (8-11). This study aimed to use LLMs to address the aforementioned gap in knowledge by constructing a comprehensive, updated, time-indexed reference dataset for CAPs in the EU documenting changes across successive SmPC versions.

# 2. Methods

## *2.1 Data sources*

Current and past versions of SmPCs for all CAPs (i.e., 1,513) were sourced from the European Commission's Union Register of Medicinal Products with a data lock point on 15/12/2025 (12).

*2.2 Construction of the Database and Time-Indexed Reference Dataset*

*2.2.1 Extraction of AEs and metadata*

Each SmPC was downloaded in portable document format (PDF). From each PDF, the brand name, the international non-proprietary name (INN), and the text of Section 4.8 were extracted. Section 4.8 was further processed to identify listed AEs using DeepSeek V3, which parsed the unstructured narrative text of Section 4.8 and returned a list of AEs in Comma-Separated Values (CSV) format (13). AEs were mapped to standardized Medical Dictionary of Regulatory Activities (MedDRA) terminology using a hierarchical mapping strategy. Mapping was first attempted through exact string matching against the MedDRA v28 Preferred Terms (PTs). At the time of analysis, existing AE extraction frameworks such as PVLens[15] were not readily adaptable to the SmPC document structure, necessitating the development of a parallel extraction and mapping workflow(14). Terms that were flagged as unmatched were retained in the dataset for transparency and then mapped manually. Regulatory and pharmacological information for each drug–AE pair was retrieved from the European Commission's Union Register of Medicinal Products webpage by collecting relevant metadata from the HTML/JSON source of the medicinal product webpage. For all the active CAPs, the processed database was organized into a single Excel file containing 36 variables (Table 1). These included drug identification information such as brand name, International Non-proprietary Name (INN), European Union marketing authorization number, and marketing authorization holder. For each INN, the five-level World Health Organization Anatomical Therapeutic Chemical (ATC) classification system codes were retrieved. AE–related variables included the origin of the event (e.g., baseline clinical trial information or post-marketing discovery), the reference date associated with the Pharmacovigilance Risk Assessment Committee (PRAC) meeting when available, and the date on which the AE was added to the SmPC (i.e., procedure closing date). For each mapped PT to the predicted SOC, higher levels of the MedDRA v28 hierarchy were retrieved, including High Level Term (HLT), High Level Group Term (HLGT), and System Organ Class (SOC), together with their corresponding codes and the mapping method used. Regulatory metadata were extracted from the European Commission's Union Register of Medicinal Products webpage included the procedure close date, procedure type, EMA procedure reference number, European Commission decision number and date, a link to the official regulatory document, and the approved therapeutic indications. Finally, a source file identifier (i.e., the original PDF) was retained to ensure full traceability of the documents used to construct the dataset.

Table 1. Variables.

| Category | Column Name | Description | Variable number |
|---|---|---|---|
| **1. Drug Identification** | Brand_Name | Commercial name of the medicinal product. | 1 |
| | inn | International Nonproprietary Name (active substance). | 2 |
| | Union_register_eu_num | EU marketing authorization number. | 3 |
| | Union_register_mah | Marketing Authorization Holder. | 4 |
| **2. Adverse Event Information** | LLM_extracted_AE | Adverse event text as extracted by the LLM from the SmPC. | 5 |
| | Source | Origin: "Clinical Trial (Baseline)" or "Post-Approval Discovery". | 6 |
| | Reference Date | Date of the PRAC meeting (if applicable). | 7 |
| | Date Added | Date when the Adverse Event was added to the SmPC. | 8 |
| **3. MedDRA Classification** | MedDRA_PT_Term | Preferred Term. | 9 |
| | MedDRA_PT_Code | Numerical code for the Preferred Term. | 10 |
| | MedDRA_HLT_Term | High Level Term. | 11 |
| | MedDRA_HLT_Code | Numerical code for the High Level Term. | 12 |
| | MedDRA_HLGT_Term | High Level Group Term. | 13 |
| | MedDRA_HLGT_Code | Numerical code for the High Level Group Term. | 14 |
| | MedDRA_SOC_Term | System Organ Class. | 15 |
| | MedDRA_SOC_Code | Numerical code for the System Organ Class. | 16 |
| | MedDRA_Match_Method | The specific method used for MedDRA mapping. | 17 |
| **4. ATC Classification** | ATC_Level_1_Code | Anatomical Main Group code (e.g., A). | 18 |
| | ATC_Level_1_Desc | Description for Level 1 (e.g., Alimentary tract). | 19 |
| | ATC_Level_2_Code | Therapeutic Main Group code (e.g., A10). | 20 |
| | ATC_Level_2_Desc | Description for Level 2 (e.g., Drugs used in diabetes). | 21 |
| | ATC_Level_3_Code | Therapeutic/Pharmacological Subgroup code. | 22 |
| | ATC_Level_3_Desc | Description for Level 3. | 23 |
| | ATC_Level_4_Code | Chemical/Therapeutic/Pharmacological Subgroup code. | 24 |
| | ATC_Level_4_Desc | Description for Level 4. | 25 |
| | ATC_Level_5_Code | Chemical Substance code. | 26 |
| | ATC_Level_5_Desc | Description for Level 5 (Specific substance). | 27 |
| **5. Regulatory Metadata** | Union_register_close_date | Date the SmPC procedure was closed. | 28 |
| | Union_register_procedure | Type of regulatory procedure. | 29 |
| | Union_register_Ema_number | EMA procedure reference number. | 30 |
| | Union_register_decisio_number | European Commission (EC) decision number. | 31 |
| | Union_register_decision_date | Date of the EC decision. | 32 |
| | Union_register_link | URL to the official EC document. | 33 |

|  | Union_register_indication | Approved therapeutic indication for the drug. | 34 |
|---|---|---|---|
|  | Union_register_atc | Full ATC classification provided in JSON format. | 35 |
| **6. Traceability** | Source_File | Original source file name for audit trail purposes. | 36 |

*2.2.2 Validation*

For the latest version of the SmPC, two researchers (MK and STB) performed a manual validation of all extracted information included in the time-indexed reference dataset. For AEs, INNs, and brand names, each extracted item was independently reviewed against the source SmPC. For regulatory metadata, the information was independently verified against the medicinal product webpage by examining the HTML/JSON source. Extracted information about the AEs was categorized into one of five predefined categories: 1) correct (information present in the SmPC and accurately extracted), 2) incorrect (information not present in the SmPC but erroneously extracted), 3) missing (information present in the SmPC but not extracted), 4) duplicate (information extracted twice), or 5) triplicate (information extracted three or more times).

*2.2.3 Software*

All data extraction and processing were performed using Python (version 3.2) within a Google Colab environment. The following libraries and tools were employed: *requests* for automated web requests, *pandas* for structured data manipulation, *re* and *json* for parsing embedded web content, *pandas* for data management, and *pdfplumber* for robust text extraction from PDF documents. Additional libraries including *PyMuPDF*, *PyPDF2*, and *camelot-py[cv]* were installed to ensure compatibility with various PDF formats, including those containing tables or scanned images. The data processing workflow consisted of three sequential and fully automated stages:

*1) Generation of the Union Register Link List*: A comprehensive list of direct hyperlinks to Union Register product pages was generated programmatically from the official European Commission brand index, available at the *reg_index_brand.htm* endpoint of the community register portal. The index Excel file was downloaded and parsed using *pandas*, with the header located on the third row. Rows were filtered to retain only those flagged as centrally authorised human medicinal products (category code "CH"). For each retained product, the EU marketing authorisation number — structured as EU/1/YY/NNN — was parsed using a regular expression to extract the sequential product number (NNN), which directly corresponds to the Union Register h-number. A standardised URL was then constructed for each product following the pattern https://ec.europa.eu/health/documents/community-register/html/h{NNN}.htm, pointing to that product's dedicated Union Register page. These URLs were normalised and merged with the EMA medicines report metadata, which was downloaded from the EMA open data portal, using a case-insensitive brand name join key. The final enriched dataset, containing product URLs alongside regulatory metadata (INN, ATC code, marketing authorisation holder, approval date, and therapeutic area), was generated.

2) *Web Scraping and Metadata Extraction*: Product information and regulatory procedure records were programmatically retrieved from the EMA public HTML webpages. Embedded JavaScript variables containing structured data were parsed using regular expressions and JSON decoding. Extracted metadata included: a) Product name and EU authorization number; b) INN; c) Marketing authorization holder; d) Therapeutic indication; e) ATC classification; f) Direct EMA webpage link. For each procedure, the extracted attributes comprised procedure ID, type, EMA number, decision number, and decision date.

3) *PDF Download and Section 4.8 Extraction:* SmPCs PDFs were downloaded to a temporary directory in Colab. Section 4.8 text extraction was performed using *pdfplumber*, and the section was identified in the PDF searching for "Undesirable effects," or formatting variations such as "4.8 Undesirable," "4.8. Undesirable," or "4.8\nUndesirable." Robust logging captured extraction successes, failures, and warnings for transparency.

4) *AE Extraction and Structured Data Output*: Extracted Section 4.8 texts were processed using an API (DeepSeek V3), which identified individual AE in the text files. Duplicates and explanatory annotations were removed to produce clean, structured lists of AEs. These outputs were then merged with the product and procedure metadata and exported as CSV and Excel files, preserving all relevant identifiers and decision details. Timestamps and standardized filenames ensured reproducibility and traceability across all workflow steps.

5) *MedDRA v28 Enrichment via SOC-Based Filtering and Exact Matching:* All unique AE terms extracted across the dataset were systematically encoded against the MedDRA version 28, loaded from its official ASCII distribution files (.asc format). The full MedDRA hierarchy was reconstructed in memory by parsing the PT, HLT, HLGT, and SOC files, along with the inter-level linkage tables, to build both a top-down and a reverse predicted SOC-to-PT mapping. In the mapping stage, each unique extracted AEs undergoes a two-pass procedure. In the first pass, a case-insensitive exact string match is attempted against the full MedDRA v28 PT dictionary; AEs resolving at this stage are labelled "Exact Match" and proceed directly to hierarchy annotation. AEs that fail exact matching enter a second pass in which the LLM is queried to predict the one to three most likely SOCs from the 27 available MedDRA SOCs. These predicted SOCs serve exclusively as a search heuristic to reduce the PT candidate space; they are not retained as annotations in the final dataset. The filtered candidate pool, comprising only PTs belonging to the predicted SOCs, is then re-submitted to the language model to identify the single best-matching PT, which is validated against the in-memory PT-name-to-code dictionary and labelled "SOC-Filtered

Match." Once a PT is confirmed, the full MedDRA hierarchy is resolved deterministically by traversing the linkage tables (PT→HLT→HLGT→SOC) loaded from the official MedDRA ASCII distribution files. Because MedDRA is a multi-axial taxonomy in which a given PT may be linked to more than one SOC through distinct HLT pathways, the traversal implicitly resolves multiaxiality through the structure of Python dictionaries: the linkage tables are loaded using dict(zip(...)), which retains only one entry per key, so when a PT appears under multiple HLT pathways, only the last-encountered mapping survives in memory. Since MedDRA's ASCII distribution files list the primary SOC relationship before secondary ones, the surviving entry corresponds in practice to the secondary path as written in the file; however, this resolution is an implicit consequence of file ordering rather than an explicit primary-flag lookup, meaning the pipeline does not actively select the MedDRA-designated primary SOC but instead discards all but one path as a byproduct of dictionary construction. A single SOC is therefore assigned to each drug–AE pair, consistent with the requirement for unambiguous downstream aggregation, though users should be aware that the assigned SOC reflects the last-loaded linkage entry rather than a formally verified primary classification. It is important to note that the mapping success rate reflects technical resolution of terms within the dictionary and does not guarantee semantic precision in every case: near-match assignments — where an extracted term is mapped to the closest available PT rather than a definitively equivalent one — may introduce ambiguity not captured by aggregate accuracy metrics, and downstream users applying this dataset for signal detection should therefore treat PT- and SOC-level associations with appropriate caution, recognising that a subset of entries may carry representational uncertainty that could affect analyses relying on precise terminological specificity.

All intermediate and final outputs for the three steps were organized into a hierarchical folder structure, including raw .txt files for Section 4.8, extracted adverse events in .csv format, and consolidated metadata-rich Excel files. The workflow is fully reproducible, as all code, package versions, data extraction rules, and transformations are documented in the following Github repository: https://github.com/mauriziosessaku/UnionRegister_SmPCs_Database.

*2.2.4 Statistical Analysis*

Longitudinal growth of the database was assessed by calculating the cumulative number of drug-AEs associations from 1995 through 2025, while temporal trends in safety label updates were analyzed by tracking the annual frequency of new AE additions to the SmPC. To differentiate between safety profiles established during initial authorization and those evolving post-approval, we stratified drug-AEs by discovery phase and specifically in pre-marketing (clinical trials) or post-marketing (surveillance), and compared their proportional representation across

MedDRA SOC. We quantified the coverage of MedDRA SOCs by absolute frequency and by unique PTs according to their ubiquity across the product population, ranging from product-specific (unique) to non-specific (ubiquitous) terms. The density of safety information per medicinal product was summarized using central tendency and dispersion metrics; given the characteristic right-skew observed in the counts of unique PTs and distinct SOCs per drug, results were expressed as medians and means to fully capture the distribution's tail. Comparative analyses across therapeutic areas were conducted by grouping drug-AEs according to the WHO ATC level 1 hierarchy. Furthermore, a mechanism-based comparison was implemented to assess systematic differences in safety profiles between small molecules and biological or targeted therapies, utilizing percentage-based distribution across common SOCs to normalize for the varying sizes of the two drug-type populations. Boxplots and a Kaplan-Meier estimator for time-to-first SmPC update (survival analysis) were presented. All analyses were conducted in Python (version 3.2) and R (version RStudio 2026.01.0+392 "Apple Blossom" Release 49fbea7a09a468fc4d1993ca376fd5b971cb58e3, 2026-01-04 for windows).

*2.2.5 LLM Configuration and inference protocol*

Automated extraction of AEs was performed using the DeepSeek-V3 LLM, a Mixture-of-Experts (MoE) architecture accessed via the official DeepSeek API (Version: deepseek-chat). To ensure deterministic and reproducible outputs required for clinical reporting, the model's temperature was set to 0.0, effectively eliminating stochastic sampling and ensuring the most probable token selection. The system prompt was configured to define the model's persona as an "Expert assistant for structured data extraction," providing a task-specific constraint to minimize hallucinations. All inferences were conducted using a consistent payload structure without penalty parameters (frequency or presence), ensuring that the model's pre-trained alignment for precision and truthfulness remained the primary driver for data extraction from the source text. The model comprises a total of 671 billion parameters, utilizing a highly sparse activation strategy where only 37 billion parameters are active per token during inference. This architecture incorporates Multi-head Latent Attention (MLA) to optimize KV cache efficiency and an auxiliary-loss-free load balancing strategy to ensure specialized expert utilization without performance degradation. For this implementation, the model was accessed via the deepseek-chat API endpoint with a 128K token context window. Post-training alignment was achieved through a multi-stage process involving Supervised Fine-Tuning (SFT) on 1.5 million samples and Reinforcement Learning (RL), specifically incorporating reasoning patterns distilled from the DeepSeek-R1 series (14b), to enhance instruction-following and factuality in structured data extraction. In supplementary table 1 we provided the Tripod LLM checklist for transparency and reproducibility.

## 3. Results

A database was constructed with individual folders for each CAP. Each folder included a 'latest' subfolder containing the most recent SmPC, a text file of the extracted Section 4.8, and a CSV file of the parsed AEs. Furthermore, an 'updates' subfolder was generated to house the extracted Section 4.8 and corresponding CSV files for all historical versions of the SmPC. A processed version of the database was also compiled for all active CAPs, excluding withdrawn or discontinued products, formatted according to the specifications in Section 2.2.1 and Table 1.

### 3.1 Characterisation of the database

In total, 1,513 CAPs were available in the European Commission's Union Register of Medicinal Products at the data lock point on 15/12/2025. Between the first SmPC documented in the European Commission's Union Register of Medicinal Products and the cut-off date (26/10/1995-15/12/2025), a total of 17,763 SmPC versions were identified. The database had a temporal range of 30 years, comprising 125,026 drug-event associations.

### 3.2 Characterisation of the processed database

In the processed database, 1,479 active CAPs were available in the European Commission's Union Register of Medicinal Products at the data lock point on 15/12/2025. Between the first SmPC documented in the European Commission's Union Register of Medicinal Products and the cut-off date (26/10/1995-15/12/2025), a total of 110,823 drug-event associations were identified. The processed dataset had a temporal range of 30 years. The dataset includes a small number of drug-AE pairs in 1995, corresponding to the earliest centrally authorised products, and grew continuously throughout the observation period, reaching a total of 110,823 associations by the 15th of December 2025 (Figure 1 – Panel A). Examination of post-approval safety additions over time revealed that the annual rate of AEs inclusions into the SmPC following marketing authorization rose gradually throughout the late 1990s and 2000s, reaching an overall peak in 2012 (year when the PRAC was established) before stabilizing at a lower but sustained level in subsequent years (Figure 1 – Panel B). The added AEs to the SmPCs in the processed database were associated with a range of regulatory procedure types reflecting the full lifecycle of CAPs. Type I and type II variations, the primary regulatory mechanisms through which post-approval safety updates are incorporated into the SmPC (Figure 1 – Panel C). The Kaplan-Meier survival analysis characterizing the time from initial marketing authorization to first post-approval SmPC update revealed a rapid evolution of the safety profile in the post-marketing setting across CAP. The survival curve, expressing the probability of a product remaining without any post-approval SmPC modification as a function of time since authorization, declined sharply during the first 2 years following marketing authorization, reflecting the high frequency and early timing of initial post-approval safety updates across the product population. The median time to first post-approval SmPC update, defined as the point at which 50% of products in the dataset had received at least one post-approval modification, was approximately 2 years from the date of initial marketing authorization (Figure 1 – Panel D). The longitudinal analysis of the reference dataset reveals a dynamic evolution in the number of new types of AEs included in the SmPCs for the CAPs (Figure 2 – Panel A). A distinct "safety fingerprint" emerges between biologic/targeted therapies and small molecules across the top 12 SOCs. Small molecules demonstrate a higher proportional burden in Gastrointestinal (11.3%) and Nervous system disorders (10.0%), as well as a starkly higher relative frequency of Psychiatric disorders compared to biologics. In contrast, biologic and targeted therapies are characterized by a significantly higher prevalence of Infections and infestations and Skin and subcutaneous tissue disorders, likely reflecting their specific immunomodulatory and targeted pharmacological profiles. The comparison between small molecules and biological or targeted therapies revealed systematic co-representation of both small molecules and biological/target therapies in SOC distribution between the two drug type categories (Figure 2 – Panel B). The distribution of drug-AEs pairs across WHO ATC Level 1 anatomical groups confirmed that the dataset encompasses a broad range of therapeutic areas, though the coverage is not uniform across classes. Antineoplastic and immunomodulating agents constituted by far the dominant therapeutic class, accounting for 45,692 associations, representing 45.3% of the total dataset, reflecting the large number of centrally authorised oncology and immunology products and their characteristically complex and extensive safety profiles (Figure 3 – Panels A & D). Drug-AE pairs were broadly distributed across MedDRA SOCs. Gastrointestinal disorders constituted the most represented SOC with 11,618 associations (10.9%), closely followed by skin and subcutaneous tissue disorders (9,926 9.0%), nervous system disorders (9,901; 9.3%). Similar distribution of the drug-AE pairs was identified during both the pre-marketing (82,534, 74.5%) and post-marketing period (28,289, 25.5%) (Figure 3 – Panel B). The analysis revealed that the largest proportion of PTs were shared between many distinct medicinal products, while 21.7% of all unique PTs were associated with only a single product, representing highly product-specific ADR terms that may be of value for targeted signal detection analyses. A smaller proportion of terms were classified as ubiquitous and are likely to reflect non-specific class effects or background medical conditions rather than product-specific safety signals (Figure 3 – Panel C). The per-drug AEs analysis demonstrated a right-skewed distribution of MedDRA PT counts across the product population, with the highest density of products falling in the range of 25 to 100 PTs per drug, a mean of 68 PTs, and a median of 48 PTs. A long right tail in the distribution indicated that a small number of products were associated with several hundred distinct ADR terms (Figure 4 – Panel A). The median number of distinct MedDRA SOCs associated with individual medicinal products across the full dataset was 14, indicating that most centrally authorized products are associated with ADRs spanning multiple organ systems simultaneously. Antineoplastic and immunomodulating agents and nervous system drugs displayed the highest median SOC counts and the widest interquartile ranges, reflecting the systemic and pleiotropic nature of their pharmacological and toxicological profiles (Figure 4 – Panel B).

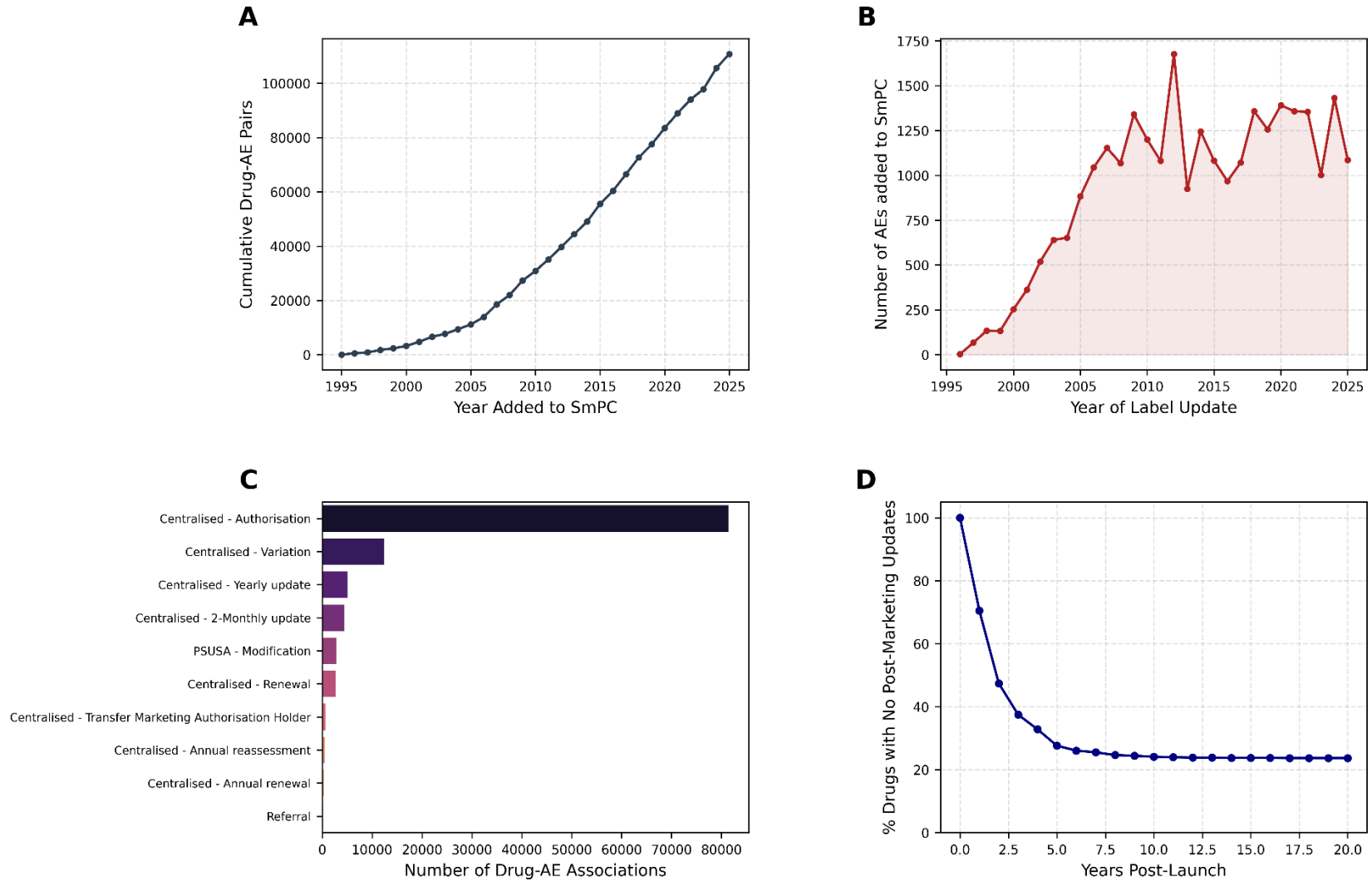

**Figure 1.** Characterization of the time-indexed reference dataset – part 1. <u>*Legend*</u>: *Panel A) Longitudinal growth of the time-indexed reference dataset*; Panel B) Number of AEs added to SmPCs; *Panel C) Regulatory procedures; Panel D) Percentage of drugs with post-marketing updates of the SmPCs*

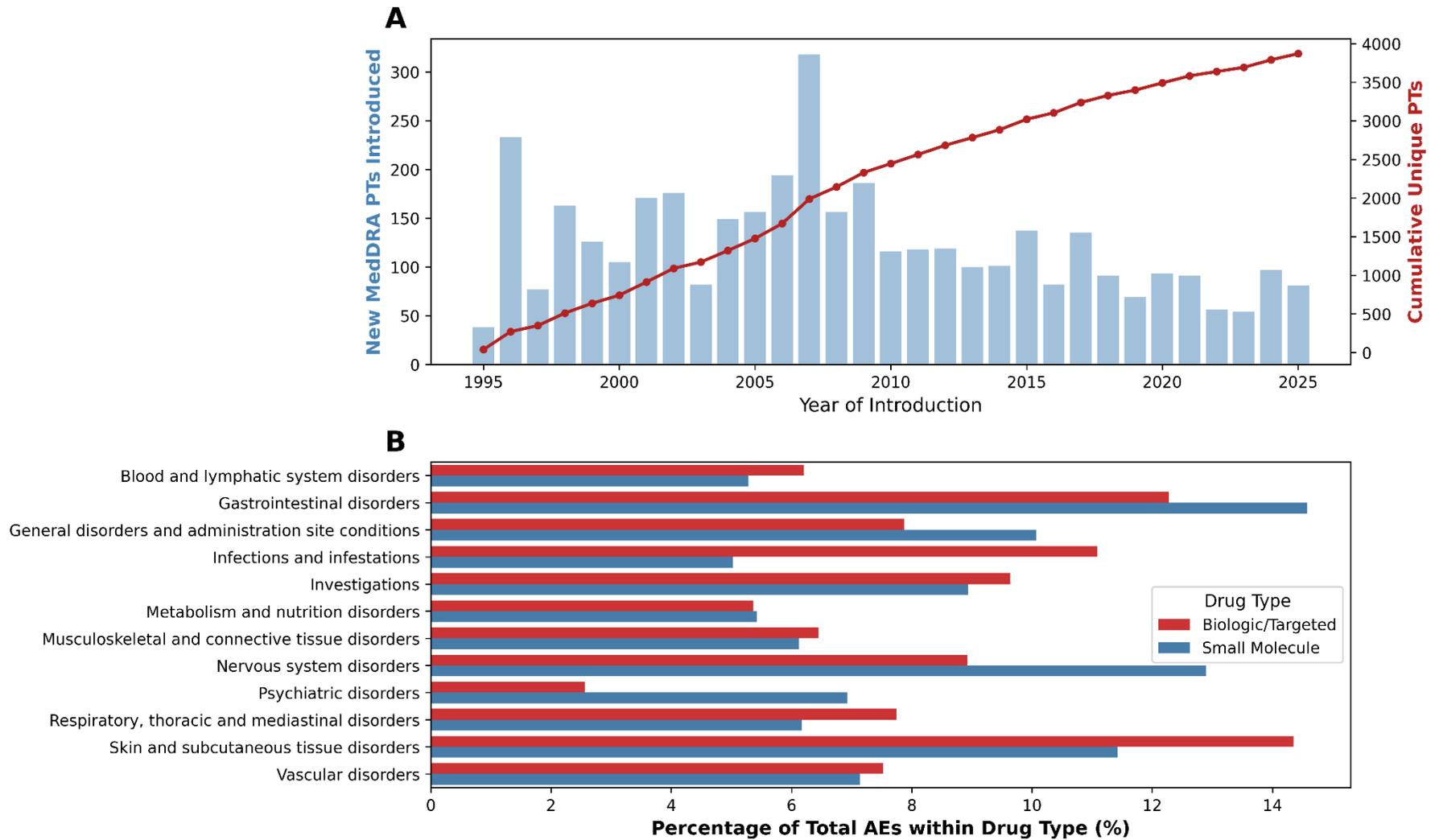

**Figure 2.** Characterization of the time-indexed reference dataset – part 3. <u>Legend</u>: *Panel A) Introduction of New PTs*; *Panel B) Biologic/targeted drugs and small molecule.*

<u>Note</u>: *Drugs were classified as Biologic/Targeted according to the classification provide in DrugBank (https://go.drugbank.com/biotech_drugs)*

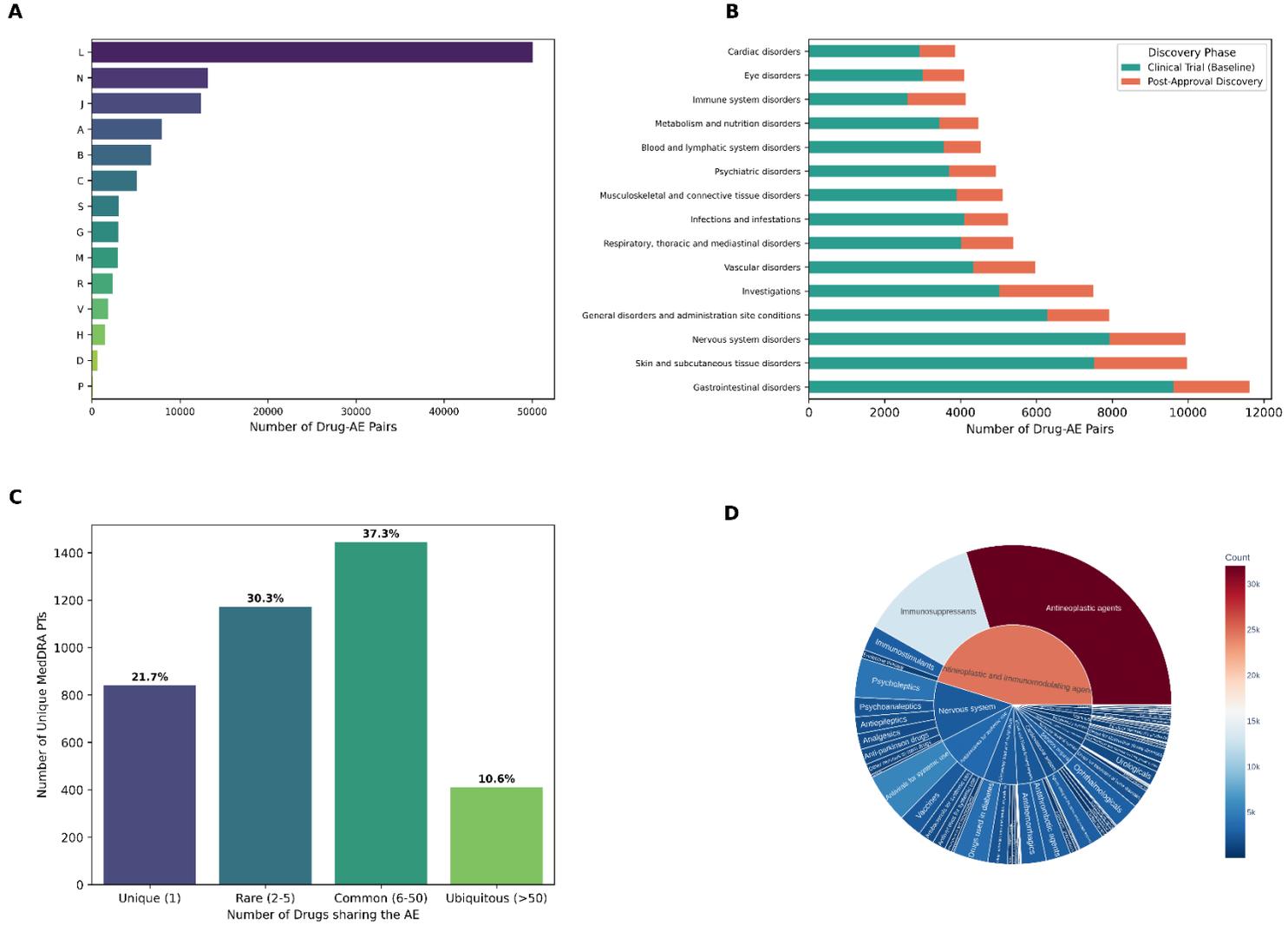

**Figure 3.** Characterization of the time-indexed reference dataset – part 2. <u>*Legend*</u>: *Panel A) ATC level 1 distribution; Panel B) Baseline vs. Post-marketing; Panel C) Uniquity; Panel D) Hierarchical AE Distribution by ATC 1st and 2nd level.*

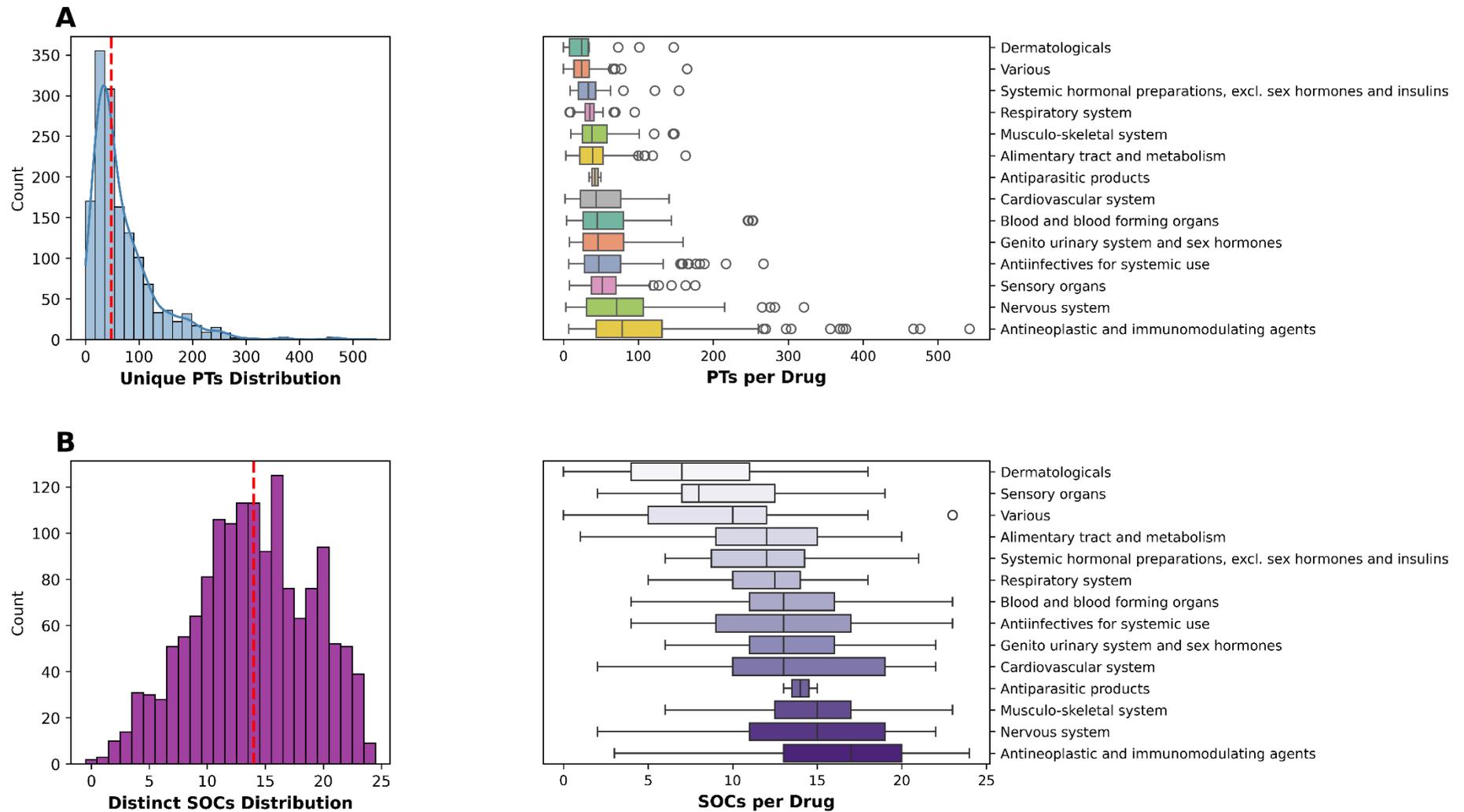

**Figure 4.** Characterization of the time-indexed reference dataset – part 4. *Legend*: *Panel A) Density and boxplot of the frequency of Unique PTs per drug; Panel B) Histogram and boxplot of the frequency of SOCs per drug.*

*3.3 Validation of the Time-Indexed Reference Dataset*

Manual validation of the DeepSeek extraction model showed 95.1% accuracy with missing events being the most prevalent error type. Missing AEs from DeepSeek extraction were manually assigned. MedDRA terminology mapping achieved an overall success rate of 95.7%, including exact string matching against the MedDRA v28 PT dictionary (73.5%) and SOC-filtered batch matching (22.2%). Following both strategies, 0.5% of terms remained unmatched and fewer than 0.01% resulted in processing errors and were manually assigned.

4. Discussion

This study developed a time-indexed reference dataset for CAPs in the EU. To our knowledge, this is the first reference dataset to simultaneously satisfy four criteria that, individually, are partially met by existing resources but have never previously been achieved in combination: 1) comprehensive population coverage of all centrally authorised medicinal products in the European Union, 2) dynamic and updatable architecture reflecting the ongoing evolution of drug safety profiles, 3) precise temporal indexing of each drug-AE association to a regulatory-grade reference date, and 4) full terminological standardisation through MedDRA and ATC annotation across the complete dataset alongside regulatory metadata. A direct comparison with the existing reference datasets most widely used in the pharmacovigilance literature (Table 2) underscores the magnitude of this advance.

**Table 2**. Existing reference datasets.

| *Dataset* | *Scale* | *Focus* | *Source(s)* | *Limitations* |
|---|---|---|---|---|
| SIDER 4.1 (Side Effect Resource) (15) | ≈140,000 drug-ADR pairs. | Marketed medicines and ADR | ADR from drug labels / package inserts | Cannot track signal emergence over time, static, outdated (2015) Never validated in any formal analysis. Does not contain vaccines or OTC. |
| Japanese drug-safety reference set 39709323 (2024) (16) | 43 drugs × 15 events; 127 positive & 386 negative controls. | safety info for drugs marketed in Japan. | Positives: J-RMP negatives: JADER | Limited to Japanese drugs and regulatory context |
| OMOP / Ryan et al. drug-outcome test set 24166222(2013)(17) | 399 test cases: 165 positive & 234 negative controls across 4 outcomes. | 4 outcomes: acute liver injury, acute kidney injury, acute myocardial | Systematic literature review + NLP of product labels to classify drug- | Evidence quality varied across outcomes. Useful as reference but not for |

| | | infarction, upper GI bleeding. | outcome pairs as positive/negative | benchmarking due to limited coverage |
|---|---|---|---|---|
| Pediatric drug-event reference set (GRiP) 25663078(2015) (18) | Reviewed 256 drug×event pairs → 37 positive (17 with pediatric evidence) & 90 negative controls (rest unclassifiable). | Pediatric drugs/events (16 drugs × 16 events chosen by expert/usage patterns) | Summary of Product Characteristics, Micromedex, literature and expert review. | Smaller, pediatric-focused set. Many pairs are unclassified. |
| PVLens arXiv: 2503.20639(2025) (19) | ≈610,000 drug-ADR pairs from all FDA SPL medicines, vaccines and OTC products (6,897 ATC substances) | Extraction of labeled safety information | FDA (SPLs) dictionary-based NLP pipeline | Currently only FDA data. Manual validation is limited to a small dataset (97 labels). |
| Pediatric vaccine ADRFI reference set 26496461(2016) (20) | 13 vaccines × 14 ADR → total 182 vaccine-ADR pairs: 18 positive, 113 negative, 51 unclassifiable. | Vaccine ADR in the pediatric population (ADRFI surveillance). | Systematic literature searches + expert reports + evidence-based resources; two independent reviewers classified pairs with consensus process. | Large fraction unclassifiable; intended for comparing signal-detection methods in pediatric vaccine safety only. |
| CRESCENDDI - DDI reference set 35246559(2022) (21) | 10,286 positive and 4,544 negative controls; covers 454 drugs and 179 ADR (mapped to RxNorm & MedDRA). | Clinically relevant adverse drug-drug interactions (DDIs), plus single-drug ADR data for included drugs. | Automated extraction from clinical resources to create a scalable reference; demonstrated by scanning spontaneous reporting system data. | Very large, DDI-focused set. Might need manual review due to noise and broad focus. |
| RS-ADR (Data-driven ADR reference standard) 36201386(2022) (22) | 1,344 drugs, 4,485 ADRs, 6,027,840 drug-ADR pairs. | Broad ADR reference set. | Integrated existing reference sets (e.g., SIDER, OMOP, EU-ADR) | Useful for large-scale validation, but evidence strength varies. Not good if you need a smaller, highly reliable benchmark. |

The OMOP/Ryan reference set, which served as one of the primary benchmarking standards in the IMI PROTECT project and remains among the most frequently cited resources in the field, comprises 399 drug-outcome combinations restricted to four clinical outcomes, acute liver injury, acute kidney injury, myocardial infarction, and upper gastrointestinal bleeding. While its methodological rigour and careful curation are well recognised, its narrow clinical scope and limited size substantially constrain its suitability for generalised benchmarking across the full spectrum of pharmacovigilance signal detection activity. SIDER 4.1, despite containing approximately 140,000 drug-AE pairs and representing a large publicly available label-based reference resource, has not been updated since 2015 and therefore fails to reflect a decade of post-marketing safety profile evolution for any of the products it covers. It carries no temporal information whatsoever, rendering it entirely unsuitable for the kind of longitudinal and pre-inclusion stratified analyses proposed in this thesis. The Japanese reference set and the GRiP paediatric reference set, while valuable for their respective regional and population-specific purposes, cover only 43 and 16 medicines respectively, limiting their utility for large-scale or therapeutically diverse benchmarking exercises. The RS-ADR dataset, one of the largest integrated reference resources with over six million drug-AE pairs, achieves its scale through automated aggregation of multiple heterogeneous source databases of variable quality and evidence strength, introducing a degree of noise and inconsistency that may undermine its reliability as a gold standard for rigorous method evaluation.

The dataset constructed in this study instead, comprising 110,823 drug-AE associations and thirty years of centralised EU regulatory activity, occupies a position that no existing resource currently fills. Its coverage spans all major therapeutic areas represented in the centralised EU portfolio, with antineoplastic and immunomodulating agents, nervous system drugs, and antiinfectives collectively accounting for high proportion of the dataset while dermatological, cardiovascular, and metabolic products ensure broad pharmacological diversity. The inclusion of 28,289 post-approval discoveries, adverse events added to the SmPC after initial marketing authorisation, representing 25.5% of the total associations, provides a clinically and methodologically relevant pool of confirmed positive controls that reflect genuine post-marketing signal detection outcomes rather than pre-approval clinical trial findings alone. The quality and validity of the dataset are further supported by the rigorous validation procedure applied to the LLM-based extraction pipeline. Unlike most existing reference datasets, which rely on automated extraction methods without systematic validation against source documents, the validation conducted in this study involved a comprehensive manual review of all 1,513 products included in the study, in which every AEs extracted by the LLM model was individually compared term by term against the corresponding Section 4.8 of the source SmPC PDF. The overall extraction accuracy of 95.1%, combined with a MedDRA mapping success rate of 95.7%, of which 73.5% were resolved through exact matching and a further 22.2% through SOC-filtered batch matching, leaving only 0.5% unresolved, confirms that the dataset maintains a high and well-characterised level of terminological integrity throughout.

*4.1 Strengths and Limitations*

Prior work comparing LLM-based extraction with ontology-constrained approaches suggests that differences between methods often reflect representational level rather than failure to identify core adverse event concepts.

LLM-based workflows tend to recover surface-level expressions from text, whereas ontology-driven pipelines normalized these expressions into structured concept hierarchies. This distinction has important implications for downstream use, as it affects both interpretability and comparability of extracted safety information. While the observed extraction accuracy of 95.1% demonstrates strong performance, several limitations inherent to LLM-based approaches should be acknowledged. In this context, LLM outputs are sensitive to linguistic variation and context, which may lead to incomplete extraction of events (i.e., missed AEs) or the generation of imprecise or non-standard terms. As observed in validation, missing events constituted the most frequent error type. Additionally, certain extracted terms may not directly correspond to standardized MedDRA PTs, requiring downstream mapping that can introduce further ambiguity or misclassification. For example, semantically related but non-identical phrases may be mapped to incorrect PTs when contextual nuance is insufficiently captured. Second, the separation between extraction and terminology standardization introduces a structural dependency between the LLM and the mapping process. While the SOC-filtered matching strategy improves efficiency and mapping success rates, it relies on correct upstream extraction and accurate SOC prediction. Errors at either stage may propagate into the final dataset. PVLens MedDRA mapper provides robust standardized coding, however it was not explicitly amenable to integration outside of the formal FDA Structured Product Labeling (SPL) processing pipeline. These observations suggest that LLM-based extraction, while powerful, may benefit from integration with more deterministic or domain-specific approaches. Hybrid frameworks that combine LLM-based parsing with rule-based or dictionary-driven methods may provide improved robustness, interpretability, and reproducibility. Such approaches can leverage the flexibility of LLMs in handling unstructured text while anchoring outputs to controlled vocabularies through deterministic matching and validation layers.

Future work should explore these hybrid strategies more explicitly, including systematic comparisons between LLM-based, rule-based, and combined methods across diverse regulatory text sources. This may enable further improvements in both extraction accuracy and terminological precision, while supporting scalable integration into broader PV workflows.

A further methodological consideration concerns the handling of MedDRA's multi-axial structure, whereby a given PT may belong to more than one SOCs through distinct hierarchical pathways. In the present pipeline, this is resolved implicitly rather than explicitly: the inter-level linkage tables are loaded using Python dictionary construction (dict(zip(...))), which retains only one entry per key when duplicates are encountered, discarding all but the last-loaded mapping. Because MedDRA's ASCII distribution files list the primary SOC relationship before secondary ones, the surviving entry corresponds in practice to the secondary path as written in the file order; the pipeline therefore does not actively select the MedDRA-designated primary SOC but resolves multiaxiality as a byproduct of dictionary construction and file ordering. While this approach produces consistent results across the dataset given a fixed MedDRA version, it represents an implicit dependency on the stability of the ASCII file structure across MedDRA releases, and the assigned SOC for any multi-axial PT should not be assumed to reflect the formally designated primary classification without independent verification. Users conducting SOC-level signal detection analyses should be aware that a proportion of drug–AE pairs involving multi-axial PTs may carry a SOC assignment that differs from the one that would result from an explicit primary-flag lookup, and stratified analyses across organ classes should be interpreted with this structural limitation in mind.

These considerations should be interpreted in the context of the overall strengths of the dataset described below. The main quality of the study is grounded in the structural soundness and novelty of the dataset. The dataset was constructed from authoritative data ensuring that the positive control designations to drug-AE pairs are grounded in formal regulatory decisions rather than expert opinion, literature, or automated label extraction. The use of SmPC closing dates as temporal anchors provides a level of precision and legal defensibility that cannot be secured from any alternative source. Furthermore, the dataset achieves unprecedented scale and therapeutic breadth for an EU-specific reference standard as well as high terminological integrity. It also ensures independence from the spontaneous reporting data used for signal detection, preventing circularity due to overlapping. Lastly, pre-inclusion analysis is transparently carried out and fully reproducible with all data, code, and documentation publicly available, facilitating independent validation and extension by other researchers and supporting longitudinal monitoring.

Nonetheless, other limitations warrant careful consideration. First, the reference dataset covers only CAPs, representing a subset of all medicines available in the EU. Nationally authorised products, constituting many medicines, are not included. Therefore, the findings may not be fully generalisable to analyses conducted on databases of a broader product lineup. Additionally, the SmPC regulatory closing date used as reference for pre-inclusion stratification, represents the formal update approval date rather than the date of first identification of the safety signal. The actual signal detection and validation process precedes the SmPC update by up to several months, as documented by the average signal-to-label latency.

This work is also complementary to emerging efforts aimed at large-scale extraction and standardization of regulatory safety information from drug labels. Recent frameworks such as PVLens[15] have begun to focus on the systematic processing of FDA SPLs, providing scalable processing pipelines for extracting and normalizing adverse event information across a large corpus of regulatory documents. While the present study focuses on European SmPCs and introduces temporal indexing as a key innovation, the underlying objectives are closely aligned. The dataset developed here may serve as a natural extension to such frameworks, particularly PVLens, enabling future integration of EU and US regulatory sources and supporting cross-regulatory analyses of drug safety profiles. This alignment also reinforces the potential value of hybrid extraction approaches that combine LLM-based methods with established dictionary-driven mapping processes. Beyond benchmarking applications, this dataset may serve as a foundational resource for downstream PV systems requiring temporally resolved, regulator-aligned safety information. While the present study focuses on dataset construction and characterization, future work will be required to operationalize this resource within signal detection frameworks.

## 5. Conclusion

In this study, the first comprehensive, time-indexed reference dataset covering all centrally authorised medicinal products in the EU was successfully curated, documenting thirty years of regulatory activity with precise timestamps, MedDRA and ATC annotation. This dataset will open the opportunity for strengthening benchmarking strategies for signal detection in the EU.

## 6. Data Availability

All the code and data utilized for this work are openly available from the following Github: https://github.com/mauriziosessaku/UnionRegister_SmPCs_Database.

## 7. Conflict of interests

None

## 8. Author´s contribution

Conceptualization: MS, JLP; Methodology: MS, STB, MK, JLP; Software: STB, MS, JLP; Validation: MK, STB, MS; Formal Analysis: MS, STB, MK; Investigation: MK, JLP, STB, MS; Resources: MS; Data Curation: STB, MS, MK; Writing – Original Draft Preparation: MK, JLP, STB, MS; Writing – Review & Editing: MK, JLP, STB, MS; Visualization: STB, MK; Supervision: MS; Project Administration: MS.

# SUPLEMENTARY MATERIAL

**Supplementary Table 1.** TRIPOD-LLM Checklist.

| Section | Item | Description | Page no. |
|---|---|---|---|
| **Title** | 1 | Identify the study as developing, fine-tuning and/or evaluating the performance of an LLM, specifying the task, the target population and the outcome to be predicted. | 1 |
| **Introduction** | | | |
| **Abstract** | 2 | See TRIPOD-LLM for abstracts. | 2 |
| **Background** | 3a | Explain the healthcare context/use case (for example, administrative, diagnostic, therapeutic and clinical workflow) and rationale for developing or evaluating the LLM, including references to existing approaches and models. | 3 |
| | 3b | Describe the target population and the intended use of the LLM in the context of the care pathway, including its intended users in current gold standard practices (for example, healthcare professionals, patients, public or administrators). | N/A |
| **Objectives** | 4 | Specify the study objectives, including whether the study describes the initial development, fine-tuning or validation of an LLM (or multiple stages). | 3 |
| **Methods** | | | |
| **Data** | 5a | Describe the sources of data separately for the training, tuning and/or evaluation datasets and the rationale for using these data (for example, web corpora, clinical research/trial data, EHR data or unknown). | 4 |
| | 5b | Describe the relevant data points and provide a quantitative and qualitative description of their distribution and other relevant descriptors of the dataset (for example, source, languages and countries of origin). | 4-5, 9 |
| | 5c | Specifically state the date of the oldest and newest item of text used in the development process (training, fine-tuning and reward modeling) and the evaluation datasets. | 4, 9 |
| | 5d | Describe any data preprocessing and quality checking, including whether this was similar across text corpora, institutions and relevant sociodemographic groups. | 4-7 |
| | 5e | Describe how missing and imbalanced data were handled and provide reasons for omitting any data. | 6, 9-10 |
| **Analytical Methods** | 6a | Report the LLM name, version and last date of training. | 4 |
| | 6b | Report details of the LLM development process, such as LLM architecture, training, fine-tuning procedures and alignment strategy (for example, reinforcement learning and direct preference optimization) and alignment goals (for example, helpfulness, honesty and harmlessness). | 4 |
| | 6c | Report details of how the text was generated using the LLM, including any prompt engineering (including consistency of outputs), and inference settings (for example, seed, temperature, max token length and penalties), as relevant. | 4-7 |
| | 6d | Specify the initial and postprocessed output of the LLM (for example, probabilities, classification and unstructured text). | 4-7 |
| | 6e | Provide details and rationale for any classification and, if applicable, how the probabilities were determined and thresholds identified. | N/A |
| **LLM Output** | 7a | Include metrics that capture the quality of generative outputs, such as consistency, relevance, accuracy and presence/type of errors compared to gold standards. | 6, 16 |
| | 7b | Report the outcome metrics' relevance to the downstream task at deployment time and, where applicable, the correlation of metric to human evaluation of the text for the intended use. | 6, 16 |
| | 7c | Clearly define the outcome, how the LLM predictions were calculated (for example, formula, code, object and API), the date of inference for closed-source LLMs and evaluation metrics. | 4-7 |
| | 7d | If outcome assessment requires subjective interpretation, describe the qualifications of the assessors, any instructions provided, relevant information on demographics of the assessors and inter-assessor agreement. | 6 |
| | 7e | Specify how performance was compared to other LLMs, humans and other benchmarks or standards. | N/A |
| **Annotation** | 8a | If annotation was done, report how the text was labeled, including providing specific annotation guidelines with examples. | 4-6 |
| | 8b | If annotation was done, report how many annotators labeled the dataset(s), including the proportion of data in each dataset that was annotated by more than one annotator, and the inter-annotator agreement. | 4-6 |
| | 8c | If annotation was done, provide information on the background and experience of the annotators or the characteristics of any models involved in labeling. | N/A |
| **Prompting** | 9a | If research involved prompting LLMs, provide details on the processes used during prompt design, curation and selection. | N/A |
| | 9b | If research involved prompting LLMs, report what data were used to develop the prompts. | N/A |
| **Summarization** | 10 | Describe any preprocessing of the data before summarization. | N/A |
| **Instruction tuning/ alignment** | 11 | If instruction tuning/alignment strategies were used, what were the instructions, data and interface used for evaluation, and what were the characteristics of the populations doing the evaluation? | N/A |
| **Compute** | 12 | Report compute, or proxies thereof (for example, time on what and how many machines, cost on what and how many machines, inference time, floating-point operations per second), required to carry out methods. | 6-8 |
| **Ethical approval** | 13 | Name the institutional research board or ethics committee that approved the study and describe the participant-informed consent or the ethics committee waiver of informed consent. | N/A |
| **Open science** | 14a | Give the source of funding and the role of the funders for the present study. | N/A |
| | 14b | Declare any conflicts of interest and financial disclosures for all authors. | 6 |
| | 14c | Indicate where the study protocol can be accessed or state that a protocol was not prepared. | N/A |
| | 14d | Provide registration information for the study, including register name and registration number, or state that the study was not registered. | N/A |
| | 14e | Provide details of the availability of the study data. | 20 |
| | 14f | Provide details of the availability of the code to reproduce the study results. | 20 |
| **Public involvement** | 15 | Provide details of any patient and public involvement during the design, conduct, reporting, interpretation or dissemination of the study or state no involvement. | N/A |
| **Results** | | | |
| **Participants** | 16a | When using patient/EHR data, describe the flow of text/EHR/patient data through the study, including the number of documents/questions/participants with and without the outcome/label and follow-up time as applicable. | N/A |
| | 16b | When using patient/EHR data, report the characteristics overall and for each data source or setting and development/evaluation splits, including the key dates, key characteristics and sample size. | N/A |
| | 16c | For LLM evaluation that includes clinical outcomes, show a comparison of the distribution of important clinical variables that may be associated with the outcome between development and evaluation data, if available. | N/A |
| | 16d | When using patient/EHR data, specify the number of participants and outcome events in each analysis (for example, for LLM development, hyperparameter tuning and LLM evaluation). | N/A |
| **Performance** | 17 | Report LLM performance according to prespecified metrics (see item 7a) and/or human evaluation (see item 7d). | 16 |
| **LLM updating** | 18 | If applicable, report the results from any LLM updating, including the updated LLM and subsequent performance. | N/A |
| **Discussion** | | | |
| **Interpretation** | 19a | Give an overall interpretation of the main results, including issues of fairness in the context of the objectives and previous studies. | 16-18 |
| **Limitations** | 19b | Discuss any limitations of the study and their effects on any biases, statistical uncertainty and generalizability. | 18-20 |
| **Usability of the LLM in context** | 19c | Describe any known challenges in using data for the specified task and domain context with reference to representation, missingness, harmonization and bias. | 18-20 |
| | 19d | Define the intended use for the implementation under evaluation, including the intended input, end-user and level of autonomy/human oversight. | N/A |
| | 19e | If applicable, describe how poor quality or unavailable input data should be assessed and handled when implementing the LLM; that is, what is the usability of the LLM in the context of current clinical care. | N/A |
| | 19f | If applicable, specify whether users will be required to interact in the handling of the input data or use of the LLM, and what level of expertise is required of users. | N/A |
| | 19g | Discuss any next steps for future research, with a specific view of the applicability and generalizability of the LLM. | 18-20 |

*Legend: M = LLM methods; D = de novo LLM development; E = LLM evaluation; H = LLM evaluation in healthcare settings; C = classification; OF = outcome forecasting; QA = long-form question answering; IR = information retrieval; DG = document generation; SS = summarization and simplification; MT = machine translation; API = application programming interface.*